%% file: multiscale_meshgraphnets_main.tex
\documentclass[nohyperref]{article}

\usepackage{microtype}
\usepackage{graphicx}
\usepackage{subfigure}
\usepackage{booktabs} % for professional tables

\usepackage{hyperref}

\input{defns.tex}

\usepackage[accepted]{ai4science} 

\usepackage{amsmath}
\usepackage{amssymb}
\usepackage{mathtools}
\usepackage{amsthm}
\usepackage{dsfont}
\usepackage{tikz}
\usetikzlibrary{decorations.pathreplacing,calligraphy}

\usepackage[capitalize,noabbrev]{cleveref}

\theoremstyle{plain}

\theoremstyle{definition}

\theoremstyle{remark}

\usepackage[textsize=tiny]{todonotes}

\icmltitlerunning{MultiScale MeshGraphNets}

\begin{document}

\twocolumn[
\icmltitle{MultiScale MeshGraphNets}



\icmlsetsymbol{equal}{*}

\begin{icmlauthorlist}
\icmlauthor{Meire Fortunato}{equal,deepmind}
\icmlauthor{Tobias Pfaff}{equal,deepmind}
\icmlauthor{Peter Wirnsberger}{deepmind}
\icmlauthor{Alexander Pritzel}{deepmind}
\icmlauthor{Peter Battaglia}{deepmind}
\end{icmlauthorlist}

\icmlaffiliation{deepmind}{DeepMind}

\icmlcorrespondingauthor{Meire Fortunato}{meirefortunato@google.com}

\icmlkeywords{Graph Neural Networks, Simulations, Multiscale, Mesh, ICML}

\vskip 0.3in
]



\printAffiliationsAndNotice{\icmlEqualContribution} % otherwise use the standard text.

\begin{abstract}
In recent years, there has been a growing interest in using machine learning to overcome the high cost of numerical simulation, with some learned models achieving impressive speed-ups over classical solvers whilst maintaining accuracy. However, these methods are usually tested at low-resolution settings, and it remains to be seen whether they can scale to the costly high-resolution simulations that we ultimately want to tackle.

In this work, we propose two complementary approaches to improve the framework from MeshGraphNets, which demonstrated accurate predictions in a broad range of physical systems.
MeshGraphNets relies on a message passing graph neural network to propagate information, and this structure becomes a limiting factor for high-resolution simulations, as equally distant points in space become further apart in graph space. First, we demonstrate that it is possible to learn accurate surrogate dynamics of a high-resolution system on a much coarser mesh, both removing the message passing bottleneck and improving performance; and second, we introduce a hierarchical approach (MultiScale MeshGraphNets) which passes messages on two different resolutions (fine and coarse), significantly improving the accuracy of MeshGraphNets while requiring less computational resources. 

\end{abstract}

% Include paper content from external files 
\input{sessions/introduction}
\input{sessions/model_and_high_labels}
\input{sessions/datasets}
\input{sessions/results}

\input{sessions/related_work}
\input{sessions/conclusions}

\section*{Acknowledgements}
We would like to thank Alvaro Sanchez-Gonzalez, Yulia Rubanova, Charles Blundell, Michael P. Brenner, and our reviewers for valuable discussions on the work and feedback on manuscript.

\bibliography{references}
\bibliographystyle{icml2022}

%%%%%%%%%%%%%%%%%%%%%%%%%%%%%%%%%%%%%%%%%%%%%%%%%%%%%%%%%%%%%%%%%%%%%%%%%%%%%%%
%%%%%%%%%%%%%%%%%%%%%%%%%%%%%%%%%%%%%%%%%%%%%%%%%%%%%%%%%%%%%%%%%%%%%%%%%%%%%%%
% APPENDIX
%%%%%%%%%%%%%%%%%%%%%%%%%%%%%%%%%%%%%%%%%%%%%%%%%%%%%%%%%%%%%%%%%%%%%%%%%%%%%%%
%%%%%%%%%%%%%%%%%%%%%%%%%%%%%%%%%%%%%%%%%%%%%%%%%%%%%%%%%%%%%%%%%%%%%%%%%%%%%%%
\newpage
\appendix
\onecolumn
\input{sessions/appendix.tex}
%%%%%%%%%%%%%%%%%%%%%%%%%%%%%%%%%%%%%%%%%%%%%%%%%%%%%%%%%%%%%%%%%%%%%%%%%%%%%%%
%%%%%%%%%%%%%%%%%%%%%%%%%%%%%%%%%%%%%%%%%%%%%%%%%%%%%%%%%%%%%%%%%%%%%%%%%%%%%%%

\end{document}

%% file: defns.tex
\newcommand{\squishlist}{
   \begin{list}{$\bullet$}
    { \setlength{\itemsep}{0pt}      \setlength{\parsep}{3pt}
      \setlength{\topsep}{3pt}       \setlength{\partopsep}{0pt}
      \setlength{\leftmargin}{1.5em} \setlength{\labelwidth}{1em}
      \setlength{\labelsep}{0.5em} } }

\newcommand{\squishlisttwo}{
   \begin{list}{$\bullet$}
    { \setlength{\itemsep}{0pt}    \setlength{\parsep}{0pt}
      \setlength{\topsep}{0pt}     \setlength{\partopsep}{0pt}
      \setlength{\leftmargin}{2em} \setlength{\labelwidth}{1.5em}
      \setlength{\labelsep}{0.5em} } }

\newcommand{\squishend}{
    \end{list}  }

\newcommand{\calD}{{\cal D}}

\newcommand{\calG}{{\cal G}}

\newcommand{\multimeshnets}{\textsc{MS-MGN}}

\newcommand{\maybevspace}[1]{}

%% file: sessions/introduction.tex
\section{Introduction}

There has been a growing interest in accelerating or replacing costly traditional numerical solvers with learned simulators, which have the potential to be much faster than classical methods \citep{nils2020navierstokes,dima2021accelerated,pfaff21MeshGraphNets, keisler2022}. Furthermore, learned simulators are generally differentiable by construction, which opens up interesting avenues for inverse design \citep{challapalli2021inverse,Goodrich2021,allen2022InverseDesign}.
A recent approach to learning simulations discretized on unstructured meshes is MeshGraphNets (MGN, \citet{pfaff21MeshGraphNets}), which encodes the simulation mesh at each time step into a graph, and uses message passing Graph Neural Networks (GNNs,~\citet{gilmer2017neural, scarselli2008graph,battaglia2018relational}) to make predictions on this graph. MGN demonstrated strong generalization, and accurate predictions on a broad range of physical systems. 

The accuracy of traditional solvers is often limited by the resolution of the simulation mesh. This is particularly true for chaotic systems like fluid dynamics: processes at very small length-scales, such as turbulent mixing, affect the overall flow and need to be resolved on very fine meshes to accurately solve the underlying partial differential equation. This leads to the characteristic \emph{spatial convergence}; where simulation accuracy increases monotonically with the mesh resolution. This is an important property for the use of numerical solvers in practice, as it allows trading in compute to obtain the desired solution accuracy.

However, it is unclear whether this behavior also applies to learned simulation approaches, particularly GNN-based models like MGN. There are reasons to believe it is not the case: as the mesh becomes finer, message passing GNNs have to perform more update steps to pass information along the same physical distance. This results in significantly higher computational cost, and may also cause over-smoothing \citep{LiOversmoothing, ChenOversmoothing}. 

In this work, we investigate MGN on highly resolved meshes, and find that message propagation speed indeed becomes a limiting factor, leading to high computational costs and reduced accuracy at high resolutions. To overcome this limitation, we propose two orthogonal approaches:
\begin{itemize}
\item First, we introduce MultiScale MeshGraphNets (\multimeshnets{}), a hierarchical framework for learning mesh-based simulations using GNNs, which runs message passing at two different resolutions. Namely, we have message passing on input (fine) mesh but also at a coarser mesh that facilitates the propagation of information. We demonstrate that \multimeshnets{} restores spatial convergence, and is more accurate and computationally efficient than MGN.
\item Second, we modify the training distribution to use high-accuracy labels that better capture the true dynamics of the physical system. 
As opposed to simply replicating the spatial convergence curve of traditional solvers, this allows to make \emph{better} predictions than the reference simulator at a given resolution.
\end{itemize}
Together, these approaches are a key step forward for learned mesh-based simulations, and improve accuracy for highly resolved simulations at a lower computational cost. 

%% file: sessions/model_and_high_labels.tex
\section{MultiScale MeshGraphNets}
\label{sec:method}
Here we introduce MultiScale MeshGraphNets (\multimeshnets{}), a hierarchical version of MeshGraphNets (MGN). As in MGN, the model uses a message passing GNN to learn the temporal evolution of physical systems discretized on meshes. In contrast to MGN, passes are being made on both the graph defined by the fine input mesh, and in a coarser mesh. This coarse mesh is introduced only with the aim of promoting more efficient communication in latent space, to efficiently model fast-acting or non-local dynamics. All inputs and outputs are defined on the fine input mesh. 

This architecture is inspired by both empirical findings about message propagation in graphs and by multigrid methods \cite{briggs2000multigrid,bramble2019multigrid}. First, message propagation speed in Cartesian coordinates is bounded by the length of the mesh edges multiplied by the number of message passing blocks. Refining the mesh aiming to obtain greater precision decreases the lengths of the edges, which implies a lower speed of propagation of information. This can lead to certain effects not being modeled properly on high-resolution meshes. Using an auxiliary coarse mesh, we can retain high message propagation speeds even for very fine input meshes. And second, GNNs are related to Gauss-Seidel smoothing iterations as they can only reduce errors locally. By solving the system at multiple resolutions, multigrid methods demonstrate an effective way to achieve global solutions using local updates.

\begin{figure}[t]
	\centering
	\includegraphics[width=1.0\columnwidth]{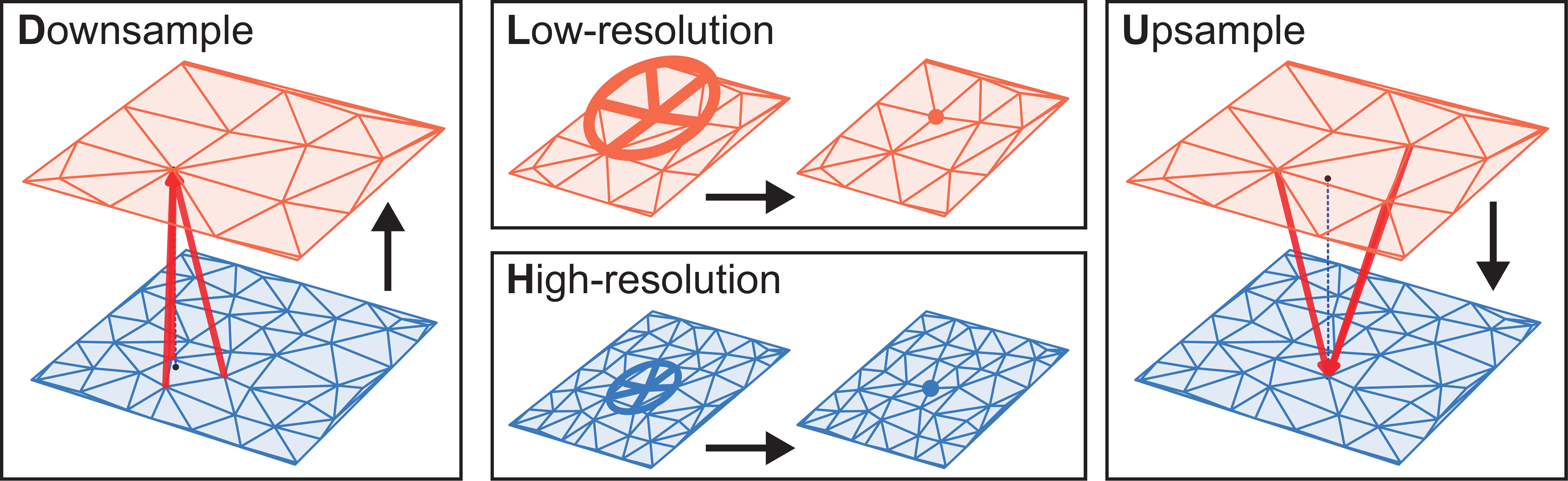}
	\caption{The four update operators on \multimeshnets{}: \textbf{D}ownsample (left), where each node on the low-resolution mesh (orange mesh) receives information from the high-resolution mesh triangle (blue mesh) enclosing the node; \textbf{H}igh-resolution (bottom-middle), where high-resolution nodes are updated by their connected neighbors; \textbf{L}ow-resolution (top-middle), where low-resolution nodes are updated by connected their neighbors; \textbf{U}psample (right), where each high-resolution node receives information from the corresponding low-resolution nodes it updates in the \textbf{D}ownsample update.}
	\label{fig:down_up_sample}
\end{figure}

\multimeshnets{} uses the Encode-Process-Decode GNN framework introduced in \citet{sanchez2020learning}, and is trained for next-step predictions and applied iteratively to unroll trajectories at inference time. For training, encoding and decoding, we closely follow the MGN architecture. In this work, we focus on Eulerian dynamics, hence we only need to consider mesh edges and can omit world edges. The algorithm is described for 2D triangular meshes but it can also be applied for e.g. hexahedral or tetrahedral meshes.
In departure from MGN, messages are passed independently on two graphs, the coarse graph $\calG^l$ and fine graph $\calG^h$. Additionally, we define the upsampling and downsampling graphs $\calG^{up}$, $\calG^{down}$ to propagate information between levels. The training loss is only placed on nodes of the fine input graph $\calG^h$. Below, we describe graph construction and message passing operators for these graphs. The four graph operators are visualized in Figure~\ref{fig:down_up_sample}, a more detailed description of encoding and message passing can be found in the Appendix (\ref{sec:architecture_details}).

\paragraph{Encoder} A mesh is an undirected graph $\calG=(V, E)$ specified by its nodes $V$ and edges $E$. Let $\calD \subset \mathds{R}^2$ be the physical domain where the problem is defined and let $\calG^h=(V^h, E^h)$ and $\calG^l=(V^l, E^l)$ denote high-resolution and low-resolution mesh representations of $\calD$, respectively.

We encode the fine input graph $\calG^h$ as in \citet{pfaff21MeshGraphNets}, with the same node and edge features, and identical latent sizes of 128. The coarse graph $\calG^l$ is encoded in a similar fashion. However, we only encode geometric features in the coarse graph, i.e., relative node coordinates on edges, and a node type to distinguish between internal and boundary nodes. The input field variables such as velocity are only encoded into $\calG^h$.

We next construct the downsampling graph $\calG^\text{down}=(V^{l} \cup V^h, E^{h,l})$ as follows: For each fine-mesh node $i \in V^h$ we find the triangle on the coarse mesh which contains this node. Then, we create three edges $\mathbf{k}^{h,l}: i \to j$ which connect the node $i$ to each corner node $j=j(i) \in V^l$ of the triangle.\footnote{For meshes with other element types (e.g. hexahedrons or tetrahedrons) we can do the same, by finding the containing element and connecting to all corner nodes.}
As the nodes in this graph are already defined above in $\calG^h, \calG^l$, we only need to define the edge feature encoding. The edge features are the relative node coordinates from senders and receivers, which are embedded using an MLP of the same architecture as in the MGN encoders.

The upsampling graph $\calG^\text{up}=(V^{l} \cup V^h, E^{l,h})$  has the same structure. That is, for each $i\in V^l$, we create three edges $\mathbf{k}^{l,h}: i \to j$ connecting $i\in V^l$ to the corner nodes of the triangle in the high-res input mesh that contains the node $i$. Edge features are encoded exactly as in the downsampling graph. Figure~\ref{fig:down_up_sample} shows a representation for both graphs. 

\paragraph{Processor} The processor consists of several iterated processor blocks, which compute updates on the graphs defined above by message passing, as in MGN.
However, as opposed to the single graph in MGN, we now have four graphs to update. We could construct a processor block which performs updates on all four graphs simultaneously, and repeat this block $n$ times. But this would be inefficient: For example, as inputs and outputs are defined on $\calG^h$ only, the first and last updates on the other graphs would be wasted. We also note that updates on the coarse graph are significantly cheaper due to the smaller number of nodes and edges, and propagate information further. Hence, a more efficient strategy may be to perform a number of updates on $\calG^h$ to aggregate local features, downsample to the coarse graph, perform updates on $\calG^l$, upsample, and perform a few updates on $\calG^h$ to compute small-scale dynamics. This is a similar strategy as a $V$-cycle in multigrid methods; and as in multi-grid methods, we can stack several of these cycles to reduce errors even further. We will investigate a few choices for efficient processor architectures in Section \ref{sec:results}.

\paragraph{Decoder and state updater.} The next-step state predictions will be produced from the updated node latents on the fine graph $\calG^h$, and exactly follow the description in MGN, including the loss function, and hyperparemeters in the training setup.

\section{High-accuracy labels}
\label{sec:high_accuracy_labels}
The prediction quality of a learned model is bounded by the quality of the data it is trained on.

In \citet{pfaff21MeshGraphNets}, the training examples were obtained by running a traditional reference simulator for a given initial state and space discretization, and using the simulator's predictions as training labels. 
However, these labels are only approximations to the actual physical phenomena we are trying to model, and their accuracy is directly linked to the spatial resolution of the simulation. To reduce the amount of computation associated with high-resolution simulations, traditional solvers often simulate on a coarser grid, and employ heuristic ``closure models'' to approximate the effect of small-scale dynamics below the resolution of the simulation mesh \citep{pope_book}. 

In a learned model, we have another option: Instead of a handwritten heuristic, we can modify the training distribution to use ``high-accuracy labels'', e.g. by running a reference simulation at higher spatial resolution, and bi-linearly interpolating the solution down to the mesh to be used for prediction (Figure \ref{fig:high_acc_labels}). This way, the model can implicitly learn the effect of smaller scales without any changes to the model code, and at inference time potentially achieve solutions which are \emph{more accurate} than what is possible with a classical solver on a coarse scale.
In Section \ref{sec:results} we show that learning such a subgrid-aware model for fluid dynamics is indeed tractable over a surprisingly broad resolution scale.

\begin{figure}[ht]
	\centering
	\includegraphics[width=0.9\columnwidth]{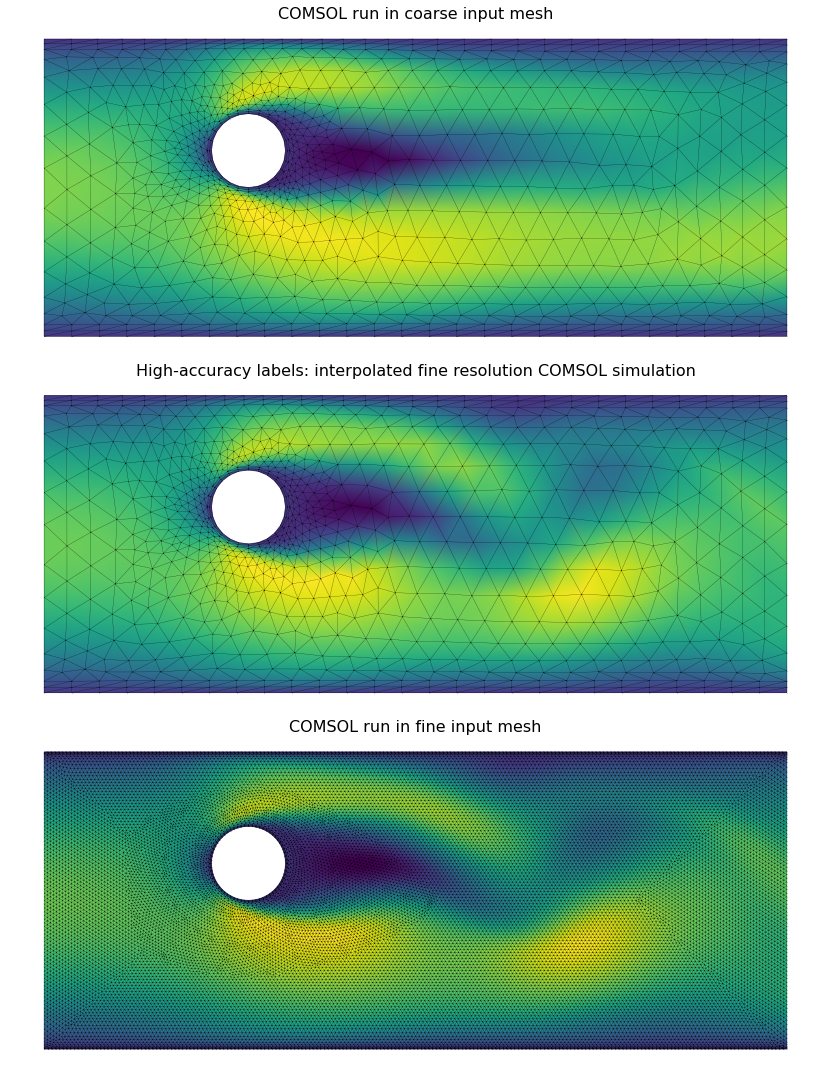}
	\caption{A reference simulation of the Karman vortex street simulated with COMSOL. The colormap shows the $x$-component of the velocity field. Top: The simulation mesh is not fine enough to resolve all flow features, and the characteristic vortex shedding is suppressed. Bottom: A more expensive simulation on a finer mesh correctly resolves the dynamics. Middle: ``high-accuracy'' labels from the high-resolution simulation (bottom) interpolated on the coarse mesh from (top), with vortex-shedding still visible. We use this to generate a training set.
}
	\label{fig:hr_labels}
\end{figure}

%% file: sessions/datasets.tex
\section{Experimental setup}
\label{sec:experimental_setup}
\paragraph{Training set}
We generated a ``CylinderFlow'' dataset comprising $1000$ trajectories of incompressible flow past a long cylinder in a channel, simulated with COMSOL~\citep{comsol}. Each trajectory consists of $T=200$ time steps, and we vary parameters, such as radius and position of the obstacle, inflow initial velocity and mesh resolution. Notably, the mesh resolution covers a wide range from a hundred to tens of thousands of mesh nodes. Section \ref{sec:dataset} in the Appendix shows example trajectories and dataset statistics. 

\paragraph{Mesh generation}
Together with the input mesh and the associated trajectories, we also use COMSOL to generate the coarse mesh $\calG^l$ used in the \multimeshnets{} architecture. To this end, we constrain the mesh generation to a specific minimum and maximum edge length (maximum edge length is always set to 5 times the minimum length). COMSOL generates adaptive meshes based on certain conditions, for example, that the mesh contains smaller edges and more nodes close to obstacles and domain boundaries.
We use the same generator to produce a high-quality mesh for both the simulation and for the coarse mesh, which conforms to the domain boundaries, and mirrors the relative node density of the simulation mesh. We use a coarse mesh of the same, fixed resolution (minimum edge length of $10^{-2}$) for all simulation meshes, to ensure sufficient message propagation speeds on all simulation meshes.

\begin{figure}[ht]
	\centering
	\includegraphics[width=1.0\columnwidth]{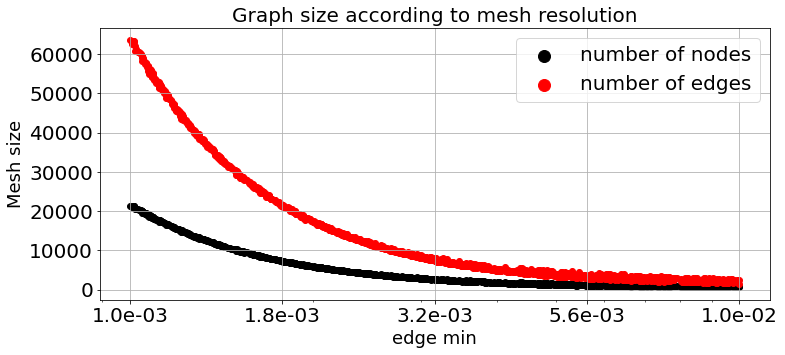}
	\vspace*{-0.5cm}
	\caption{Relationship between the minimum mesh edge length (\textit{edge min}) and the number of nodes and edges on the input mesh for the CylinderFlow test dataset.}
	\label{fig:graph_size}
\end{figure}

\paragraph{Measuring ground-truth error}
To evaluate the effects of the mesh resolution on the model's predictions we create a test dataset with 500 trajectories with varying mesh resolutions, but otherwise constant initial conditions. The minimum edge length (\textit{edge min}) of these meshes ranges from $10^{-2}$ to $10^{-3}$. Figure~\ref{fig:graph_size} shows the relationship between \textit{edge min} and node and edge count-- we note that mesh complexity increases strongly for smaller values of \textit{edge min}.

 As analytical solutions are in general not available for nontrivial simulation setups, high-resolution simulations are commonly used as a proxy for the ``ground-truth'' solution of the underlying PDE. Here, we generate such a ground-truth reference trajectory ($u_\text{ref}$) by running COMSOL at the maximum resolution in this dataset (\textit{edge min}=$10^{-3}$). In the following, we measure error by performing next-step prediction on the test set using a learned model or a classical solver \emph{at a given mesh resolution}, linearly interpolating the ground-truth trajectory onto the simulation mesh, and computing the MSE.

%% file: sessions/results.tex
% \subsection{Tables}
% Use \url{http://www.tablesgenerator.com/latex_tables#} to help make tables.https://ol.deepmind.host/project/61f131f55ff66355dd685a2e

\section{Results}
\label{sec:results}
We investigated the error behavior of MGN on simulations of various resolutions on the CylinderFlow domain (Section~\ref{sec:experimental_setup}). Our main finding is that message passing speed becomes a bottleneck for MGN performance for high-resolution meshes; but it is possible to lift this bottleneck using multiscale methods, and even show better one-step predictions than classical solvers at a given resolution using high-accuracy labels.

\begin{figure}[ht]
	\centering
	\includegraphics[width=1.0\columnwidth]{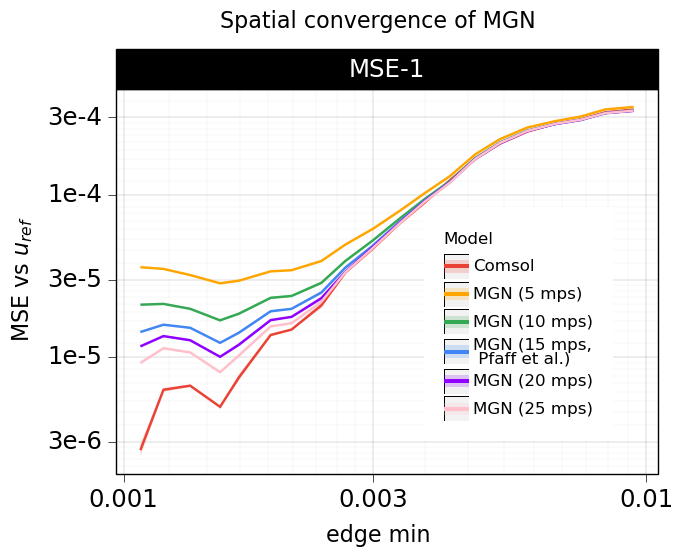}
	\vspace*{-0.5cm}
	\caption{Log-log plot of 1-step MSE measured against a highly-resolved ``ground-truth'' simulation. We expect the error to converge to 0 with decreased edge length (red curve); MGN initially tracks this curve, but accuracy drops off at small edge lengths. Increasing the number of message passing steps moves the drop-off point, but the fundamental issue remains.}
	\label{fig:edge_vs_error}
\end{figure}

\paragraph{Spatial convergence of MGN}
We first test the spatial convergence of the baseline MGN model. For this, we trained a MGN model on a mixed-resolution dataset, and tested the performance on meshes with a minimum edge length between $10^{-3}$ to $10^{-2}$. Section \ref{sec:experimental_setup} describes this setup, and how we measure error.
Figure \ref{fig:edge_vs_error} shows the spatial convergence curve of a conventional solver (red): As expected, the error decreases when moving towards finer meshes with smaller edge lengths. MGN initially tracks this behavior, but at a certain mesh resolution, the error stagnates. This is an effect of limited message propagation speed; the smaller the edge lengths become, the more message passing steps are necessary to transport information over the same physical distance. Consequently, if we increase the number of message passing steps (mps), the point of divergence from the convergence curve shifts left. However, increasing the number of message passing blocks comes at a high performance and memory cost, and the effects are diminishing; 25mps is the maximum we could train using a single GPU for this dataset, and the error still does not converge.

\paragraph{Spatial convergence of \multimeshnets{}}
Next, we compare MGN and \multimeshnets{} for different numbers of mps (15 and 25) to investigate whether multiscale message passing, as described in Section~\ref{sec:method}, can overcome the bottleneck identified for MGN in the previous paragraph. The comparison comprises two \multimeshnets{}-variants: First, a single V-cycle consisting of a single block of $\calG^h$, downsampling, 11 blocks of $\calG^l$, upsampling, and a single block of $\calG^h$, denoted as `p=1H 11L 1H (U=1,D=1)', with a total of 15 mps. Second, a variant with two V-cycles in sequence, with the architecture `p=3H 6L 3H 6L 3H (U=2, D=2)', totalling 25 mps. 

\begin{figure}[ht]
	\centering
	\includegraphics[width=1.0\columnwidth]{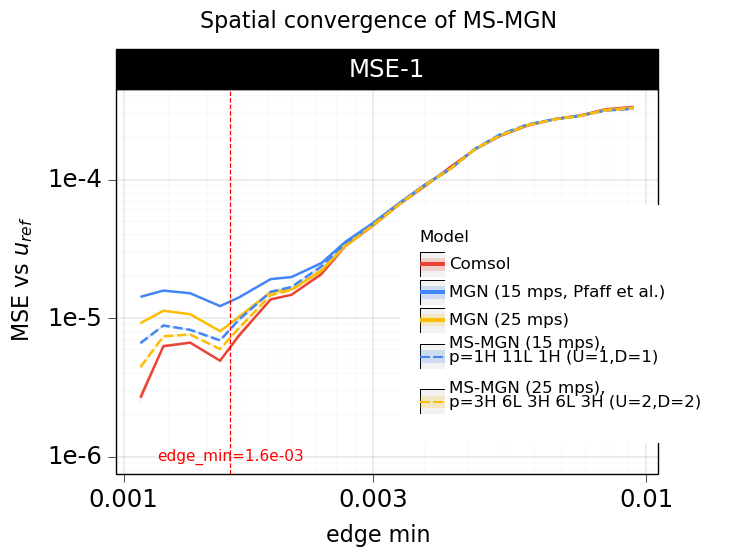}
	\vspace*{-0.5cm}
	\caption{Convergence analysis of \multimeshnets{}. With the same number of message passing steps, \multimeshnets{} exhibits a significantly lower error than the corresponding MGN model. At 25 mps, \multimeshnets{} closely tracks the convergence curve of the reference solver (red).}
	\label{fig:edge_vs_error_multi}
\end{figure}

The results in Fig.~\ref{fig:edge_vs_error_multi} show a considerable reduction in the MSE for \multimeshnets{} as compared to the MGN baseline, keeping the overall number of mps fixed. The \multimeshnets{} model with 25 mps manages to track the reference spatial convergence curve closely. This suggests that our proposed multiscale approach with the selected processor is indeed effective at resolving the message passing bottleneck for the underlying problem, and can achieve higher accuracy with the same number of mps.
In Appendix \ref{sec:addiotional_results}, we present further results to analyze the spatial variation of the error signal within the domain, which is not captured by the MSE. In particular, we perform a Graph Fourier Transform of the error signal (see, for example, \citet{Shuman2013}) and compare the results for MGN and \multimeshnets{} across the entire spectrum of eigenvalues. Our main finding of this analysis is that, for rollouts across 10 steps, the multiscale model is not only beneficial in terms of reducing the overall MSE, but that it also dampens the error across a range of eigenvalues that correspond to slowly varying (long-range) spatial correlations.

\begin{figure}[ht]
	\centering
	\includegraphics[width=0.8\columnwidth]{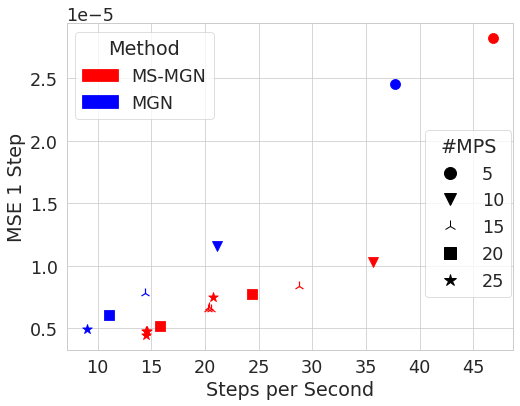}
	\vspace*{-0.5cm}
	\caption{Solution accuracy vs. training speed. \multimeshnets{} allows better performance/accuracy tradeoffs at all mesh resolutions.}
	\label{fig:accuracy_vs_time}
\end{figure}

\paragraph{Computational efficiency}
While the previous paragraph has shown that \multimeshnets{} is more accurate than MGN with the same amount of message passing steps, this is not yet the full picture. A message passing step on the low-resolution level $\calG^l$ is also computationally much cheaper than a step on $\calG^h$, as the number of node and edge updates to be computed is lower for $\calG^l$. Figure \ref{fig:accuracy_vs_time} shows the combined effect of computational cost versus accuracy in training at various mesh resolutions; the same relationship also applies at inference time. \multimeshnets{} thus allows significantly better performance/accuracy tradeoffs than the baseline. These effects are on top of the performance benefits for MGN over traditional solvers \citep{pfaff21MeshGraphNets}.

\begin{figure}[ht]
	\centering
	\includegraphics[width=1.0\columnwidth]{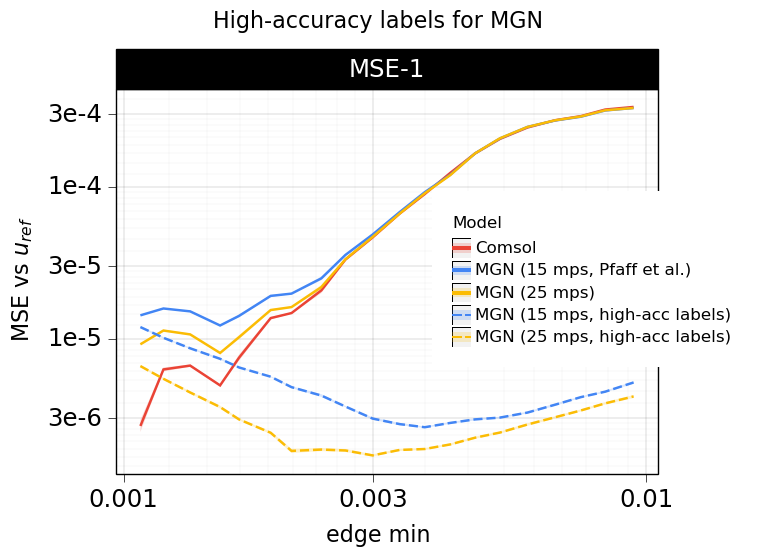}
	\vspace{-0.5cm}
	\caption{MGN baseline model trained with high-accuracy labels. At edge lengths above $0.0016$, the model is more accurate than the reference solver at a given resolution.}
	\label{fig:high_acc_labels}
\end{figure}

\paragraph{High-accuracy labels} 
Next, we train both MGN and \multimeshnets{} models on a dataset with mixed mesh resolution, but with high-accuracy labels as described in Section~\ref{sec:high_accuracy_labels}. Figure \ref{fig:high_acc_labels} shows that instead of tracking the convergence curve, the error of the MGN is actually much lower than the reference simulator. This indicates that the learned model can learn an effective model of the subgrid dynamics, and can make accurate predictions even at very coarse mesh resolutions. Surprisingly, this effect extends up to edge lengths of $10^{-2}$ which correspond to a very coarse mesh with only around a hundred nodes.

\begin{figure}[ht]
	\centering
	\includegraphics[width=1.0\columnwidth]{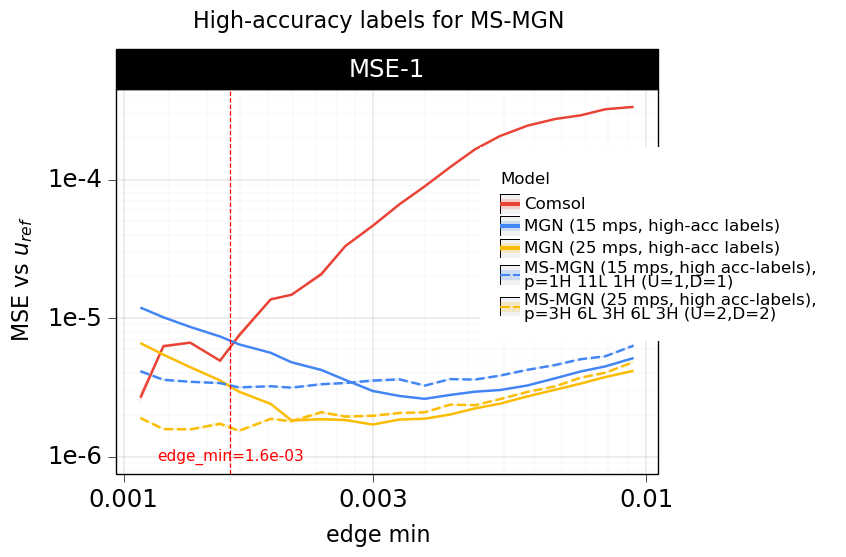}
	\vspace{-0.5cm}
	\caption{\multimeshnets{} model trained with high-accuracy labels. At 25 mps, the model is more accurate than the reference across all mesh resolutions.}
	\label{fig:edge_vs_error_hr}
\end{figure}

However, this method does not alleviate the message propagation bottleneck, and errors increase above the convergence curve for edge lengths below 0.0016. Thus, if a highly resolved output mesh is desired, accuracy is still limited. For a method that performs well both on low- and very high-resolution meshes, we can train a \multimeshnets{} model with high-accuracy labels (Figure~\ref{fig:edge_vs_error_hr}). For the 25 mps model variant, the error stays below the reference solver curve at all resolutions, with all the performance benefits of \multimeshnets{}.

\begin{figure}[ht]
	\centering
	\includegraphics[width=1.0\columnwidth]{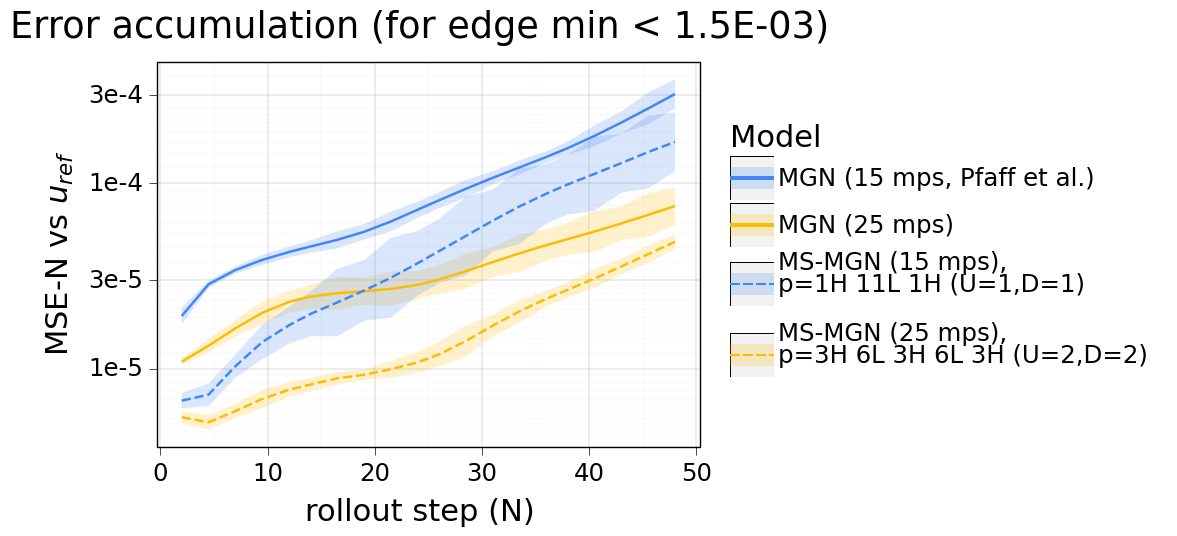}
	\vspace*{-0.5cm}
	\caption{Rollout error for simulation trajectories on high-resolution meshes.} %Note that the error accumulation of the reference solver is not measured here.}
	\label{fig:error_evolution}
\end{figure}

\paragraph{Error accumulation}
So far, we have analyzed one-step prediction error, and shown that a learned model can not only be more efficient, but reach higher accuracy at a given mesh resolution than a classical solver. However, for transient simulation, the model needs to be rolled out over many time steps, and error may accumulate. Hence, accurate one-step predictions are a necessary, but not sufficient condition for achieving accurate simulation rollouts.
While \citet{pfaff21MeshGraphNets} showed that MGN model rollouts remain stable and plausible, they can drift from the ground-truth solution. This effect is particularly pronounced for high-resolution meshes, as seen in Figure~\ref{fig:error_evolution} for meshes with edge length below 0.0015. Here, \multimeshnets{} shows reduced error compared to the baseline, but does not stop error accumulation.

\begin{figure}[ht]
	\centering
	\includegraphics[width=1.0\columnwidth]{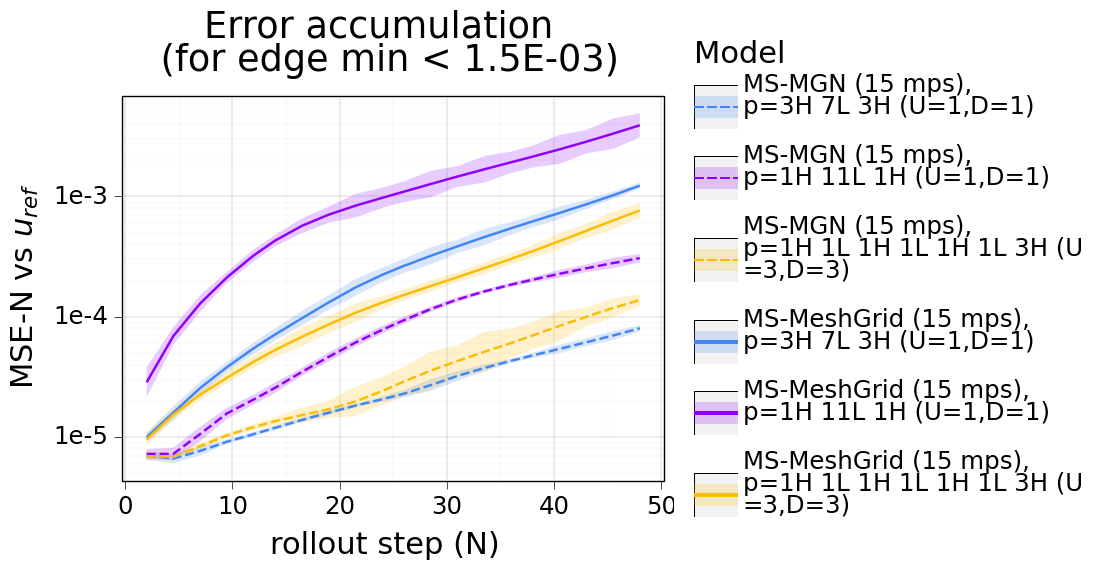}
	\vspace*{-0.5cm}
	\caption{Choice of $\calG^l$. Uniform grids show higher error than unstructured meshes created from the same distribution as the $\calG^h$.}
	\label{fig:mesh_vs_grid}
\end{figure}

\paragraph{Choice of coarse mesh}
Our default strategy for generating the coarse-level mesh $\calG^l$ is using the same mesh generator as for the simulation mesh $\calG^h$, constrained to a larger minimum edge length. However, as the method introduced in Section \ref{sec:method} is flexible in regards to mesh type, other options are possible as well. Here, we compare to a strategy similar to the one used in \citet{lino2021multignn}, which samples the simulation domain using a uniform grid. Up- and downsampling operators now operate on grid cells (creating 4 edges for downsampling), and edges ending outside the simulation domain (e.g. if they fall inside the cylinder obstacle) are omitted. Otherwise the method stays unchanged. 
Figure~\ref{fig:mesh_vs_grid} shows a comparison for both types of coarse mesh, with the grid version performing strictly worse both on 1-step ($N=1$) and rollout ($N>1$) error. We hypothesize that this is due to the fact that our default strategy produces meshes which conform to the domain boundaries, as well as retaining a similar relative node density as the simulation mesh, which makes it easier to learn universal local rules for transferring information.

%% file: sessions/related_work.tex
\section{Related Work}

\paragraph{GNN-based simulation models}
Predicting physical dynamics with GNN models is an active and growing area of research in machine learning. For example, GNNs have been successfully used for predicting dynamics of liquids and soft bodies \citep{sanchez2020learning,li2019propagation}, aerodynamics \citep{belbute2020combining}, forecasting global weather \citep{keisler2022} and quantum many-body problems \citep{kochkov2021learning}. While this work specifically studies and extends MeshGraphNets \citep{pfaff21MeshGraphNets}, we expect the broad findings to transfer to other GNN-based models, particularly those operating on meshes.

\paragraph{High-accuracy labels}
Grid-based learned simulator models sometimes operate on simulation data interpolated from a higher-resolution source \citep{thurey2020deep, dima2021accelerated}. This might be unavoidable, as the simulation output might not be in grid-format, or too fine to be processed by a CNN. However, most works do not study whether and to which degree label resolution impacts performance. A notable exception is \citet{stachenfeldCoarseMoldels} which predicts turbulent dynamics on grids using Dilated CNNs \citep{yu2016dilatedconv}, and investigates the effect of varying grid resolution. On the other hand, mesh-based GNN simulators tend to operate directly on the simulation graph. As GNN models act locally and subgrid dynamics might be harder to learn on unstructured meshes with variable local node density, it is not clear whether these models can learn effective closures and make use of high-accuracy labels. Our results suggest that this is in fact the case, and using high-accuracy labels MeshGraphNet-style models can make accurate predictions over a surprisingly broad range of mesh resolutions, allowing inference at significantly lower cost.

\paragraph{Hierarchical GNN models}
The idea of using hierarchies to structure interactions, and connect distant regions in the graph is a well-explored topic in the GNN literature, albeit in different forms and for different aims, such as representation learning~\citep{Stachenfeld2021SpectralGNN,diffpool} or geometry processing~\citep{meshcnn}.

In GNNs for physics simulation, \cite{mrowca2018flexible} use a tree hierarchy to structure particle-particle interactions in a softbody GNN physics simulation, to promote cohesion of solid objects. \citet{han2022predicting} uses a two-level hierarchy to reduce the number of nodes in a mesh-based fluid simulation, such that the system state can be fed into a transformer model without excessive memory overhead.
\citet{luz2020} use GNNs to learn restriction/prolongation operators which optimize performance of a (classical) AMG multigrid solver. And, most closely related to \multimeshnets{}, the recent work of \citet{lino2021multignn} devises a multigrid-inspired hierarchy of GNNs based on \citep{sanchez2020learning} to solve advection and incompressible flow. One key difference to \multimeshnets{} is its reliance on on-the-fly graph connectivity-- on the finest level, nodes are connected by proximity instead of mesh edge, and coarser levels use regular grids. We found this grid-based approach to perform worse than our conformal meshing, as it cannot leverage all the advantages of unstructured meshes (see Section \ref{sec:results}).

%% file: sessions/conclusions.tex
\section{Conclusions and discussion}

Message-passing GNNs can perform well and learn the dynamics of a broad range of physical problems, but for highly resolved simulations message propagation speed becomes a bottleneck, limiting the prediction accuracy. Increasing the number of message passing steps however comes with significant computational cost, and ultimately still fails to reduce error for complex meshes.

In this work, we studied this effect and proposed two complementary approaches to model global effects more efficiently when using GNNs to learn simulations. First, we demonstrated that GNN physics models can effectively leverage high-accuracy labels, to make accurate predictions even on extremely coarse simulation meshes. And second, we proposed the hierarchical model \multimeshnets{} that propagates information on two different resolutions, with a coarse discretization allowing the model to exchange information more efficiently throughout the domain. We find the proposed model to be more accurate and computationally cheaper compared to the MGN baseline with the same number of message passing steps. An interesting extension to this work would be to explore more levels in the mesh hierarchy, which will likely be beneficial for exploring even higher resolutions. Furthermore, it will be very interesting to see how our proposed multiscale approach performs on other challenging problems than the one considered in this work, for example, problems that involve more complex geometries or dynamics.

Another important research direction is reducing error accumulation for long rollouts. While rollouts remain stable for hundreds of steps for both the MGN and \multimeshnets{} models, and the multiscale approach substantially ameliorates error accumulation -- the problem is not resolved. Currently, both models are trained on next-step prediction, and unrolled with a fixed time step. Allowing for adaptive time stepping, investigating sequence models, and the role of training noise in error accumulations are exciting topics to be explored in the future.

%% file: sessions/appendix.tex
\section{Appendix}
\label{sec:appendix}

\setcounter{figure}{0}
\renewcommand{\thefigure}{A.\arabic{figure}}
\renewcommand{\theHfigure}{A.\arabic{figure}}% for hyperref

\subsection{\label{sec:mgn_details} MultiScale MeshGraphNets architecture details}
\label{sec:architecture_details}

We encode the graph $\calG^h=(V^h, E^h)$ exactly as in \citet{pfaff21MeshGraphNets}, with the same node and edge features, as well as the latent sizes (128). Each node $i \in V^h$ has embedding $\mathbf{v}_i^h$, and for each edge with sender $i\in V^h$ and receiver $j\in V^h$, we denote its edge embedding by $\mathbf{e}^h_{ij}$. Similarly,  $\mathbf{v}_i^l$ and $\mathbf{e}^l_{ij}$ denote the node and edge latents in the $\calG^l=(V^l, E^l)$ graph.

In the downsampling $\calG^\text{down}=(V^{l} \cup V^h, E^{h,l})$ graph, for each fine-mesh node $i \in V^h$, we first find the triangle on the coarse mesh that contains this node. We then create three edges $\mathbf{k}^{h,l}: i \to j$ $i\in V^h$ which connect the node $i$ to each corner node  $j=j(i) \in V^l$ of the triangle. The corresponding edge embeddings are denoted by $\mathbf{e}^{h,l}_{ij}$. The upsampling graph $\calG^\text{up}=(V^{l} \cup V^h, E^{l,h})$ has the same structure. For each $i\in V^l$, we create three edges $\mathbf{k}^{l,h}: i \to j$  to the corner nodes of the triangle in the high-res input mesh that contains the node $i$, and denote by $\mathbf{e}^{l,h}_{ij}$ their edge embeddings. 

Below we describe the node and edge feature update equations on the fine and coarse graphs (high- and low- resolution updates, respectively), as well as for the downsampling and upsampling graph operators. 

\paragraph{High-resolution update}
\begin{equation}
    \mathbf{e'}^{h}_{ij} \leftarrow f^{E,h}(\mathbf{e}^h_{ij}, \mathbf{v}^h_i, \mathbf{v}^h_j)\,,\quad
    %\mathbf{e'}^W_{ij} \leftarrow f^W(\mathbf{e}^W_{ij}, \mathbf{v}_i, \mathbf{v}_j)\,,\quad
    \mathbf{v'}^h_j \leftarrow f^{V,h}(\mathbf{v}^h_j, \sum_{i} \mathbf{e'}^h_{ij})
    \label{eq:highres_update}
\end{equation}

\paragraph{Low-resolution update}
\begin{equation}
    \mathbf{e'}^{l}_{ij} \leftarrow f^{E,l}(\mathbf{e}^l_{ij}, \mathbf{v}^l_i, \mathbf{v}^l_j)\,,\quad
    %\mathbf{e'}^W_{ij} \leftarrow f^W(\mathbf{e}^W_{ij}, \mathbf{v}_i, \mathbf{v}_j)\,,\quad
    \mathbf{v'}^l_j \leftarrow f^{V,l}(\mathbf{v}^l_j, \sum_{i} \mathbf{e'}^l_{ij})
\end{equation}

\paragraph{Downsampling update}

\begin{equation}
    \mathbf{e'}^{h,l}_{ij} \leftarrow f^{E,h,l}(\mathbf{e}^{h,l}_{ij}, \mathbf{v}^h_i, \mathbf{v}^l_j)\,,\quad
    \mathbf{v'}^l_j \leftarrow f^{V,h,l}(\mathbf{v}^l_j, \sum_{i} \mathbf{e'}^{h,l}_{ij})
\end{equation}

\paragraph{Upsampling update}

\begin{equation}
    \mathbf{e'}^{l,h}_{ij} \leftarrow f^{E,l,h}(\mathbf{e}^{l,h}_{ij}, \mathbf{v}^l_i, \mathbf{v}^h_j)\,,\quad
    \mathbf{v'}^h_j \leftarrow f^{V,l,h}(\mathbf{v}^h_j, \sum_{i} \mathbf{e'}^{l,h}_{ij})
    \label{eq:upsample_update}
\end{equation}

\noindent where the node and edge update functions in all graphs, $f^{E,h}$,  $f^{V,h}$, $ f^{E,l}$,  $f^{V,l}$, $f^{E,h,l}$,  $f^{V,h,l}$, $ f^{E,l,h}$,  $f^{V,l,h}$ are MLPs with residual connections, and the sum $\sum_{i}$ on the right equations is over all edges $\mathbf{k}: i \to j$.

\subsection{Additional results}
\label{sec:addiotional_results}
\paragraph{\multimeshnets{} processor analysis}

\begin{figure}[ht]
	\centering
	\includegraphics[ width=0.9\columnwidth]{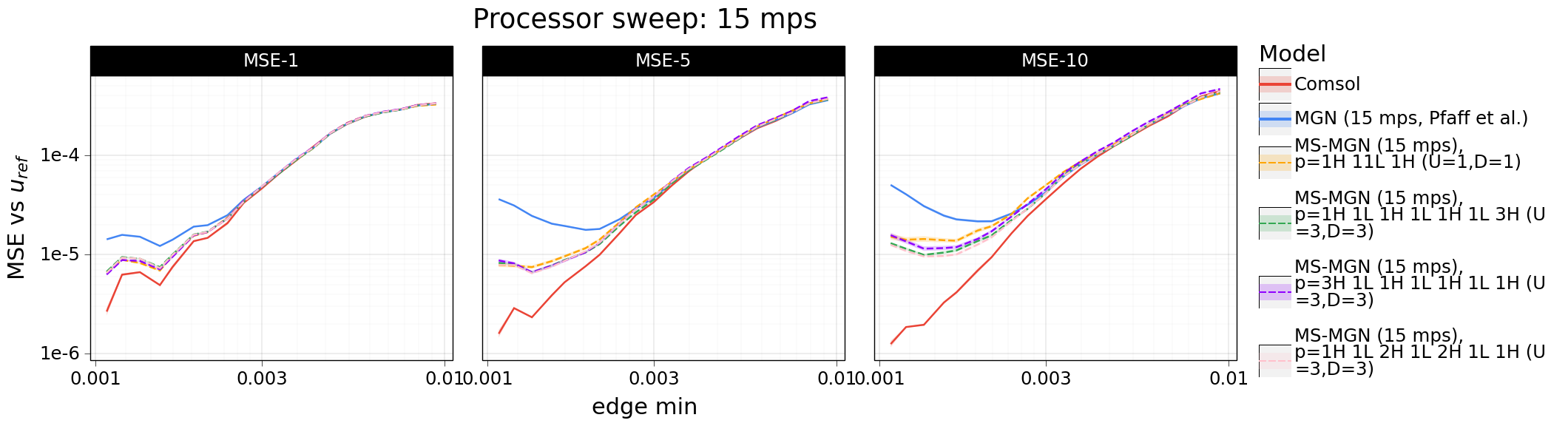}
	\includegraphics[width=0.9\columnwidth]{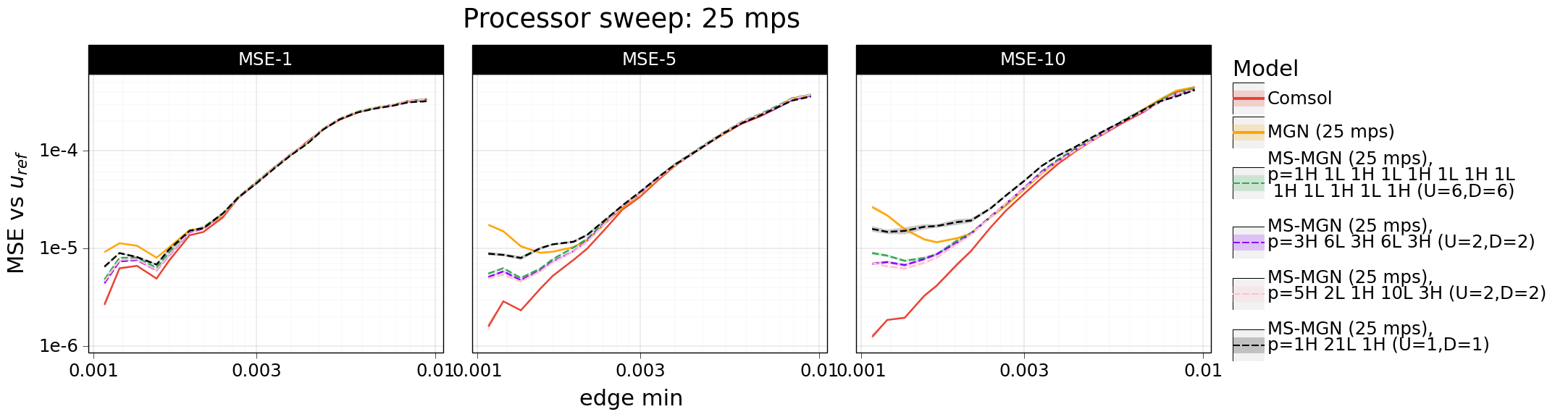}
	\caption{\label{fig:processor_sweep}\multimeshnets{} Processor parameter sweep with 5 seeds per processor. MSE-N is the average N-step mean squared error for the first N steps in the trajectory.}
\end{figure}

In figure~\ref{fig:processor_sweep}, we analyse the impact of different processor choices on the \multimeshnets{} architecture. When counting the total number of message passing steps, we consider any graph update a step, that is, we count cheaper processing steps (in the coarse mesh, or downsample and upsample graphs) in the same way as an update on the fine input mesh. We use the letters `H' and `L' to represent updates on the high- and low-resolution graphs (fine/coarse), respectively. We use letters `U' and `D' to count the total number of calls for the upsample and downsample graph operators, respectively, which happen once whenever there is a change between the fine and coarse graphs. For instance, processor `p=1H 11L 1H (U=1,D=1)' has a total of 15 steps, with 1 processing step on the fine mesh, followed by downsampling, 11 updates on the coarse level, upsampling, and a final processing step on the fine graph.

We observe that both the 1-step errors and the accumulated errors for longer rollouts are similar for the choices analysed, except for the last processor `p=1H 21L 1H (U=1,D=1)' with 25 mps, which had significantly lower accuracy. Note that its performance is similar to `p=1H 11L 1H (U=1,D=1)' with 15 mps, which indicates that no further steps are needed in the coarse mesh.  We have not investigated this in further detail but comparison with the other processors suggests that more than two message passing steps on the fine level are beneficial.

\paragraph{Graph Fourier Analysis}
The convergence analysis in figure~\ref{fig:processor_sweep} shows that our multiscale approach compares favourably to the MGN baseline in terms of the mean squared error (MSE). The MSE is a useful metric to compare the overall error computed across the entire domain but it does not reveal any insight into how the error signal varies within the domain. For a more fine-grained analysis of the latter, we can leverage techniques that are commonly employed for signal processing on graphs and decompose the signal into contributions that vary across the mesh at different frequencies~(see for example~\citet{Shuman2013} for details). To this end, we computed the Graph Fourier Transform (GFT) of the velocity error and analysed its power spectrum $|GFT|^2$ as a function of the eigenvalues of the graph Laplacian, $\{{\lambda}_n\}_{n=1,\ldots, M}$, that are assumed to be ordered such that $\lambda_1 < \lambda_2 \leq \lambda_{M-1} \leq \lambda_M$, where $M$ is the number of mesh nodes. This analysis allows us to decompose the error into contributions varying with different frequencies and to compare the results for MGN and \multimeshnets{}.

\begin{figure}[ht]
    \centering
    \begin{tikzpicture} [every node/.style={inner sep=0,outer sep=-1}]
        \draw (0, 0) node[inner sep=0] {\includegraphics[width=.49\linewidth]{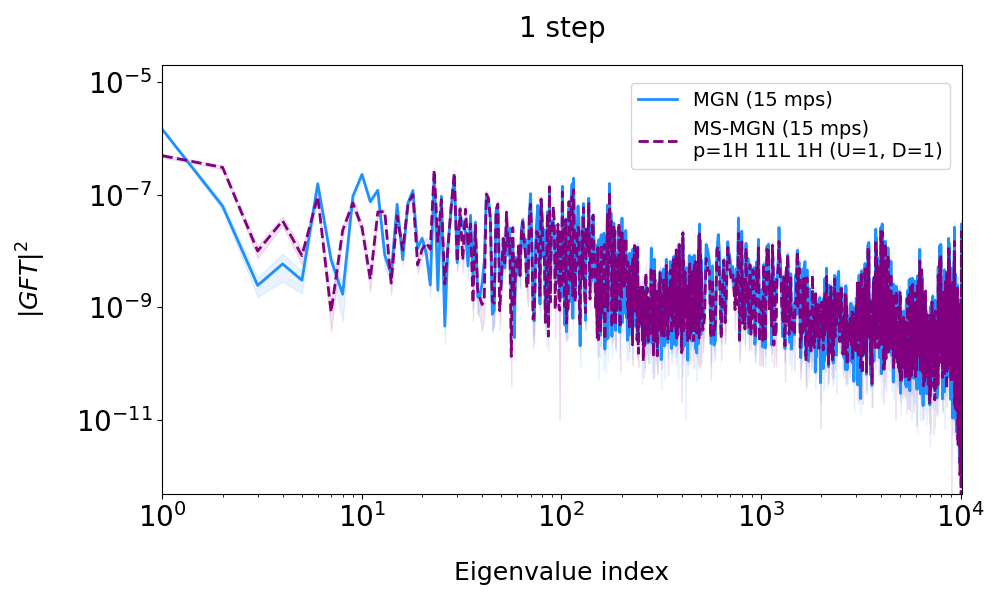}};
    \end{tikzpicture}
    \hfill
    \begin{tikzpicture} [every node/.style={inner sep=0,outer sep=-1}]
        \draw (0, 0) node[inner sep=0] {\includegraphics[width=.49\linewidth]{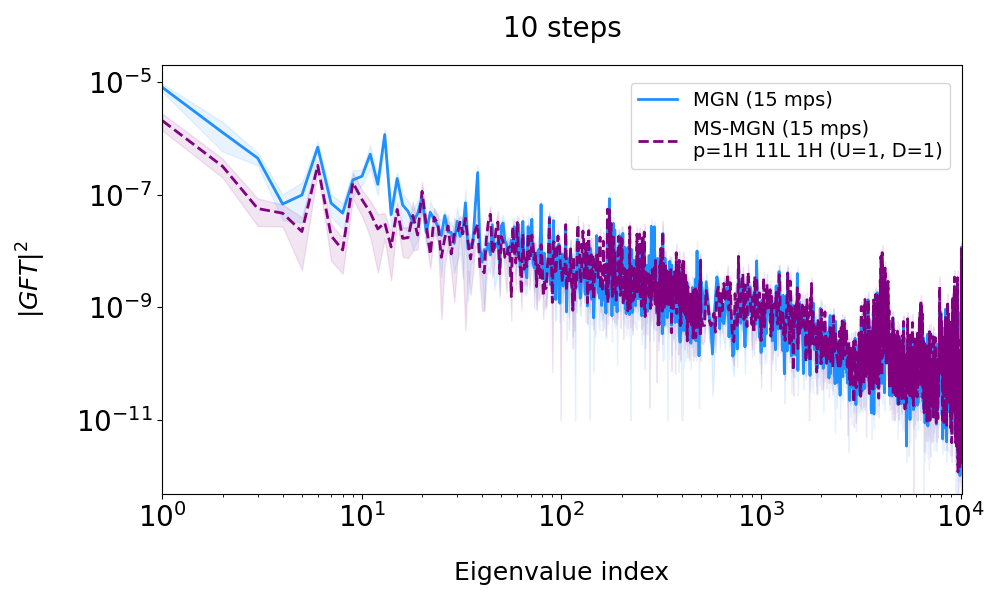}};
        \draw [decorate,decoration={brace,mirror, amplitude=5pt}]
          (-2.8,0.) -- (-0.6,0.) node[midway,  yshift=-1.5em, font=\tiny\sffamily]{\textbf{error reduction}};
    \end{tikzpicture}
	\caption{Graph Fourier analysis of the errors between the model prediction and the ground-truth velocity after one rollout step (left) and ten rollout steps (right) for the MGN baseline (solid blue) and one of the \multimeshnets{} models in Fig.~\ref{fig:processor_sweep} (dashed purple). The comparison is carried out for $\textit{edge min}= 1.5\times 10^{-3}$ and comprises 20 different seeds for each model, with shaded regions corresponding to one standard deviation of the error of the mean.}
	\label{fig:gft}
\end{figure}

\begin{figure}[ht]
	\centering
	\includegraphics[ width=1\columnwidth]{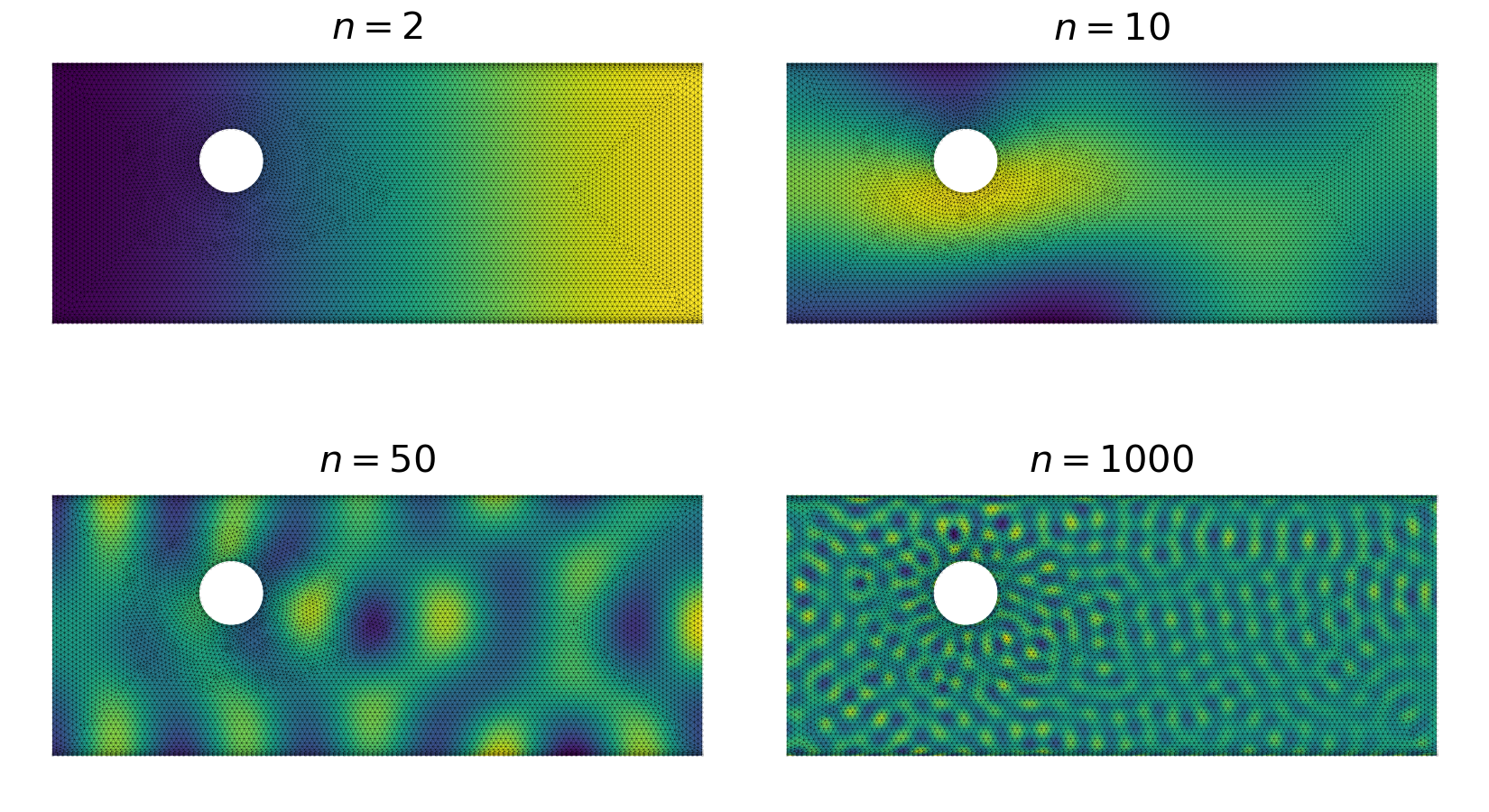}
	\caption{Illustration of the contributions of selected eigenvalues (indexed by $n$) to the absolute error in the $x$-velocity. The signal varies slowly for small $n$ (top left panel) and very fast for large $n$ (bottom right panel).}
	\label{fig:eigenvalues}
\end{figure}

Figure~\ref{fig:gft} shows the power spectrum of the error signal across the entire range of eigenvalues for a mesh with $M=10117$ nodes ($\textit{edge min}=1.5\times10^{-3}$) for next-step prediction (left panel) and for ten rollout steps (right panel). While we do not find a systematic difference between MGN and \multimeshnets{} across a single step, we observe a clear reduction in the error of the multiscale model across the first 20--30 eigenvalues for ten rollout steps. This eigenvalue range corresponds to eigenvectors that vary slowly across the mesh, as illustrated in Fig.~\ref{fig:eigenvalues}. The MSE error is captured entirely by the contribution of $\lambda_1$, whose corresponding eigenvector is a constant on the mesh. Contributions of eigenvalues with $n>1$ vary slowly across the mesh for small values of $n$ and rapidly for large values~\citep{Shuman2013}. This result therefore suggests that the \multimeshnets{} model is not only favourable in terms of reducing the MSE but that it also captures slow-varying (long-range) correlations more accurately across long rollouts, at least for the problem investigated.

% Spectrum with illustrations of individual modes.

\subsection{\label{sec:dataset}CylinderFlow Dataset}
\begin{figure}[ht]
	\centering
	\includegraphics[width=0.49\columnwidth]{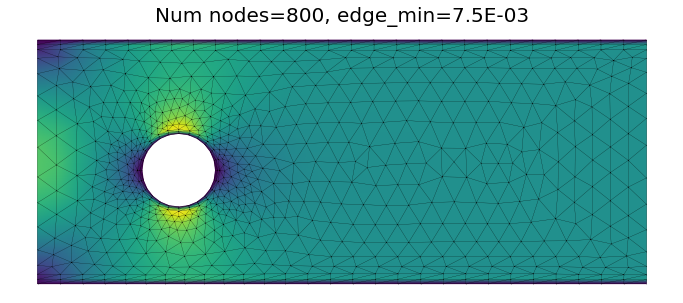}
	\includegraphics[width=0.49\columnwidth]{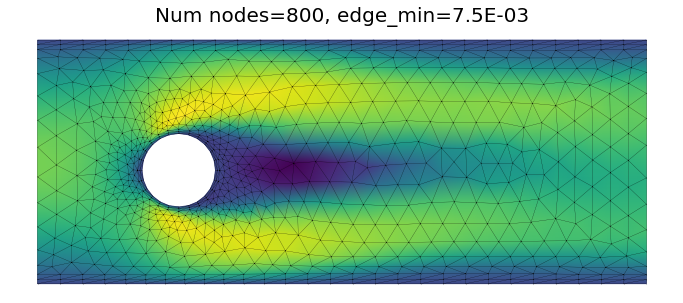}
	\includegraphics[width=0.49\columnwidth]{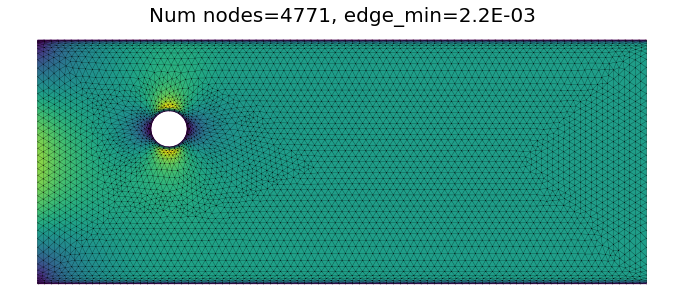}
	\includegraphics[width=0.49\columnwidth]{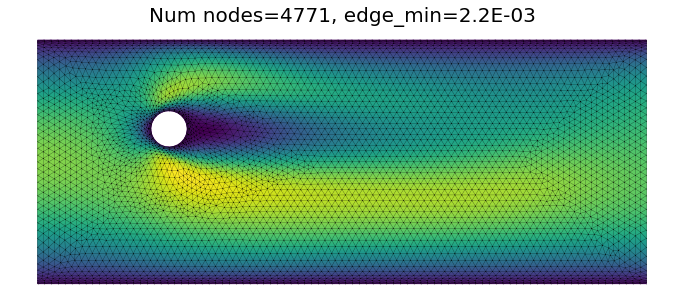}
	\includegraphics[width=0.49\columnwidth]{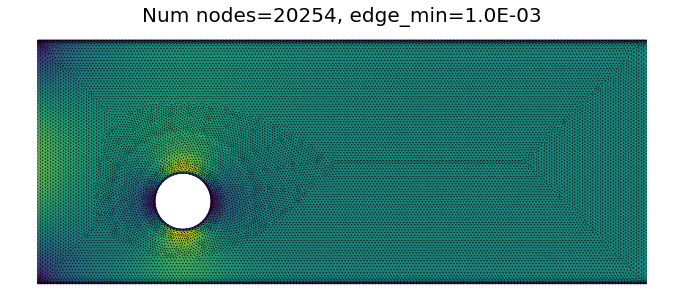}
	\includegraphics[width=0.49\columnwidth]{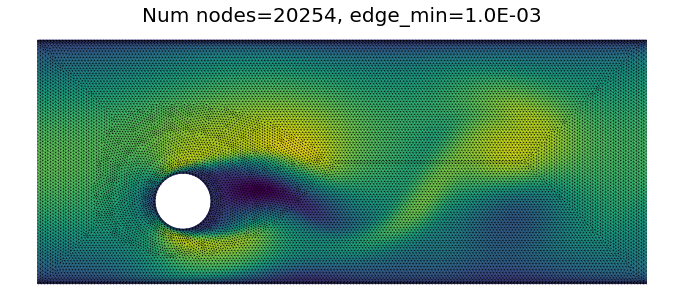}
	\caption{Initial (left) and final (right) state for different data examples on the 2D Cylinder Flow dataset. The position and size of the obstacle vary in each example, as well as the inflow initial velocity and the mesh resolution.}
	\label{fig:training_karman}
\end{figure}

We generated the 2D CylinderFlow dataset using COMSOL to simulate the temporal evolution of incompressible flow past a long cylinder in a channel. The time step is set to 0.01 seconds, and the final time on the trajectories is 2 seconds. The viscosity $\mu=10^{-3}~\text{m}^2/\text{s}$ is fixed and the height and length of the channel are set to 0.4~m and 1~m, respectively. The parameters that vary for different trajectories are the radius $R$ and center $C=(c_x, c_y)$ of the cylinder obstacle, the median initial inflow velocity $U_\text{mean}$, and the minimum edge length on the discretization mesh (\textit{edge min}). The maximum edge length is set to be 5 times the minimum one. We sampled uniformly as follows: $R \sim [0.02, 0.08]\,(\text{m})$ , $C  \sim \mathcal{U} (0.15, 0.4)  \times (0.1, 0.3) \,(\text{m})$, $U_\text{mean}\sim \mathcal{U} (0.2, 12) \,(\text{m/s}) $, $\textit{edge  min}  \sim \mathcal{L}\mathcal{U} (10^{-3}, 10^{-2})\,(\text{m})$, where $\mathcal{L}\mathcal{U}$ is the uniform distribution in log scale. In figure~\ref{fig:training_karman} we show examples of trajectories in the training set.  

\begin{figure}[ht]
	\centering
	\includegraphics[width=0.96\columnwidth]{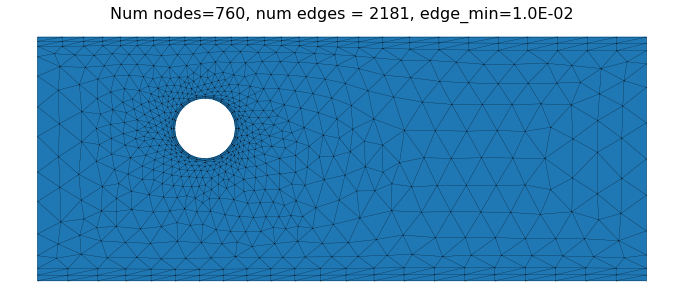}
	\includegraphics[width=0.96\columnwidth]{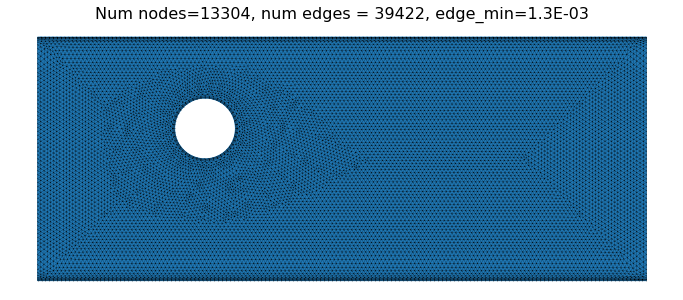}
	\includegraphics[width=0.96\columnwidth]{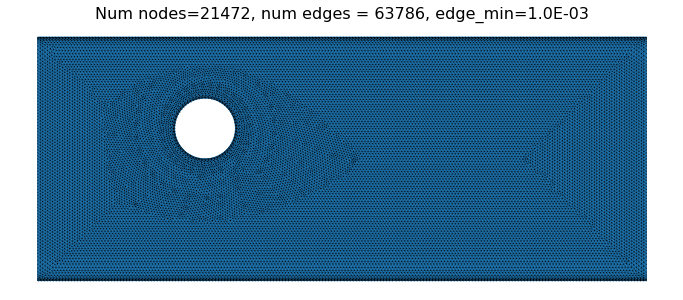}
	\caption{Different mesh resolutions in the CylinderFlow Fixed Obstacle Test Set. On top we see an example of the coarsest resolution on the test set (which is also the same resolution as the coarse mesh on the \multimeshnets{} architecture. On the bottom we see an example for the finest resolution. We note the bottom mesh has almost 30 times more nodes and edges than the top one.}
	\label{fig:mesh_resolution}
\end{figure}

In the CylinderFlow (Fixed Obstacle) test set, we fix $U_\text{mean}=0.85~\text{m/s}$, $R=0.05~\text{m}$, $C=(0.275, 0.25)~\text{m}$, and only vary the mesh resolution \text{edge  min}, sampling from the same distribution as in training. Figure~\ref{fig:mesh_resolution} shows example meshes. 

